\documentclass[conference]{IEEEtran}
\IEEEoverridecommandlockouts
\usepackage{cite}
\usepackage{amsmath,amssymb,amsfonts}
\usepackage{algorithmic}
\usepackage{graphicx}
\usepackage{textcomp}
\usepackage{xcolor}
\usepackage{graphicx}
\usepackage{amsmath}
\usepackage{caption}
\usepackage{amssymb}
\usepackage{booktabs}
\usepackage{hyperref}
\usepackage{url}
\usepackage{makecell}
\usepackage{booktabs} 
\usepackage{diagbox}
\usepackage{multirow}
\usepackage{amssymb}
\usepackage{pifont}
\usepackage{float}
\usepackage [english]{babel}
\usepackage [autostyle, english = american]{csquotes}
\MakeOuterQuote{"}
\usepackage[normalem]{ulem}
\usepackage{geometry}
\def\BibTeX{{\rm B\kern-.05em{\sc i\kern-.025em b}\kern-.08em
    T\kern-.1667em\lower.7ex\hbox{E}\kern-.125emX}}
\begin{document}

\title{CoCAtt: A Cognitive-Conditioned Driver Attention Dataset
~(Supplementary Material)}

\author{Yuan Shen, Niviru Wijayaratne*, Pranav Sriram*, \\ Aamir Hasan, Peter Du, and Katherine Driggs-Campbell

\thanks{* denotes equal contribution.}%
\thanks{Y. Shen is with the Department of Computer Science at the University of Illinois at Urbana-Champaign. N. Wijayaratne is with the Department of Mechanical Engineering at the University of Illinois at Urbana-Champaign. K. Driggs-Campbell, P. Du, P. Sriram and A. Hasan are with the Department of Electrical and Computer Engineering at the University of Illinois at Urbana-Champaign. Emails:
\{yshen47, nnw2, psriram2, aamirh2, peterdu2, krdc\}@illinois.edu}
\thanks{This work was supported by State Farm and the Illinois Center for Autonomy. This work utilizes resources supported by the National Science Foundation’s Major Research Instrumentation program, grant \#1725729, as well as the University of Illinois at Urbana-Champaign~\cite{10.1145/3311790.3396649}. }
}
\maketitle


\section{Dataset}
CoCAtt is available for download at \hyperlink{https://cocatt-dataset.github.io/}{this link}.
\subsection{Webcam Eye-tracking Data}
\begin{figure}[ht]
\centerline{\includegraphics[width=\columnwidth]{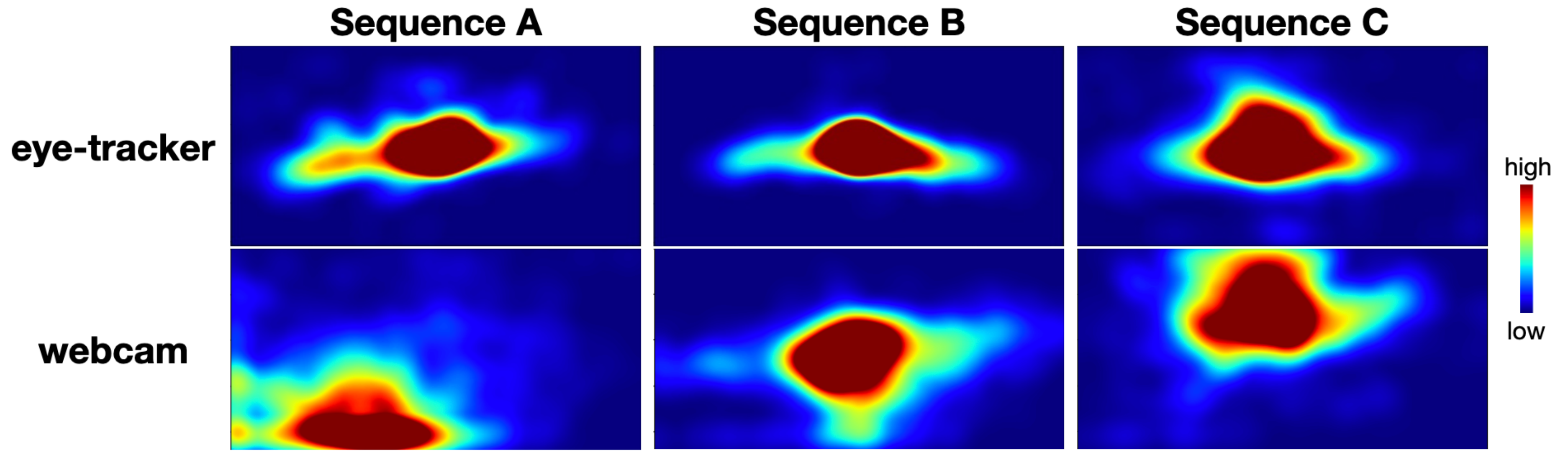}}
\caption{Cumulative heatmap for raw webcam and eye-tracker gaze captures. The three heatmaps are generated by averaging three different 10-minute raw webcam gaze captures in manual drive mode.}
\label{fig:webcam_gaze}
\end{figure}

Compared with gaze data collected by the GP3 eye-tracker, the webcam gaze shows noisier behavior in two perspectives. For illustration purposes, we randomly select three manual drive sessions from our dataset and visualize their cumulative heatmap as shown in Figure \ref{fig:webcam_gaze}. First, the noisiness can be seen from the fact that the highest density region of the cumulative attention does not always locate around the center of the driving scene, which we refer to as the center shift issue. The center shift issue is unlikely to be the actual driver attention property, since the road ahead is mostly forward, and the eye fixations should be around the scene center. Second, even if the webcam heatmap is approximately located in the center of the driving scene, the gaze data is distributed more randomly than the eye-tracker gaze. To address the noisy behavior of the webcam data, in Section 5 of our paper, we propose a simple but effective baseline architecture that uses a coarse-to-fine calibration procedure that handles the two perspectives of noisiness above respectively.

Aside from the noisy pattern of the data distribution itself, the webcam-based eye-tracking API that we use is susceptible to the presence of mask and skin color due to its dependency on facial landmark detection output. In our dataset, we mask those frames that the facial landmark detection fails due to people wearing masks. 

\section{Baseline Architectures}
\subsection{Implementation Details for the Unconditioned Baseline and Multi-branch}
We visualize our architecture for the unconditioned and multi-branch baseline models in Figure \ref{fig:multi_branch_baseline}. For the unconditioned baseline, i.e., BDD-A, we only keep one attention sub-branch, since there is no driver state as inputs. As for the multi-branch model, the number of sub-branches is the number of driver states. We select the corresponding sub-branch for each sequence based on the driver state. Note that we divide each sequence during training so that it only contains the same driver state for all frames. 

\subsection{Implementation Details for the Modified CondConv Architecture}
\begin{figure}[t]
\centerline{\includegraphics[width=0.5\textwidth]{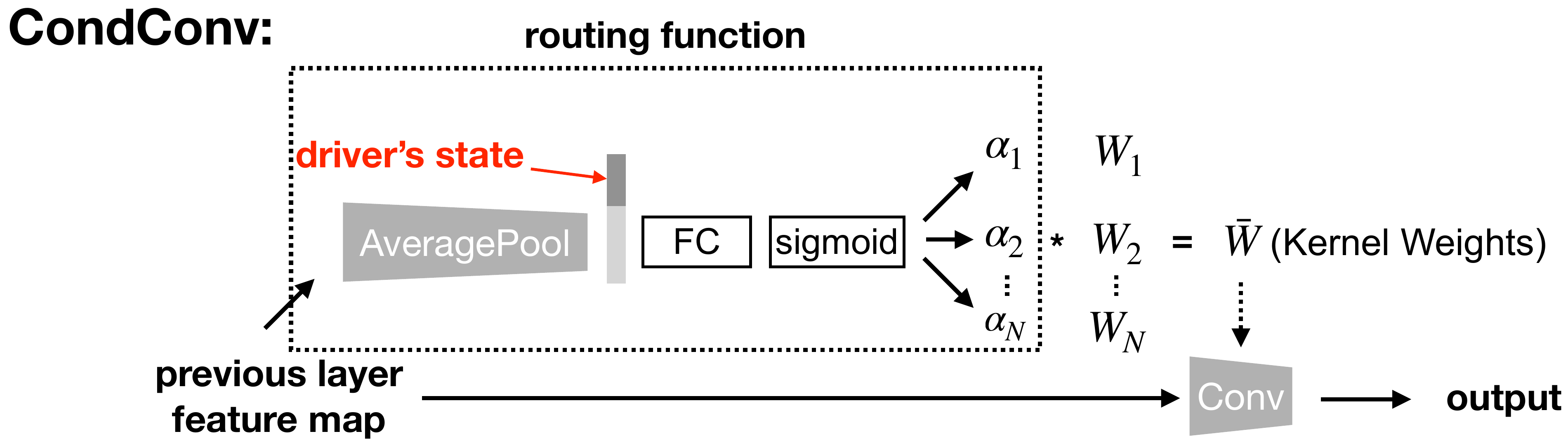}}
\caption{The architecture of the modified CondConv. In our implementation, we use four expert weights, i.e., $N=4$, within the routing functions.  \label{fig:modified_condconv}}
\end{figure}

\begin{figure}[t]
\centerline{\includegraphics[width=\columnwidth]{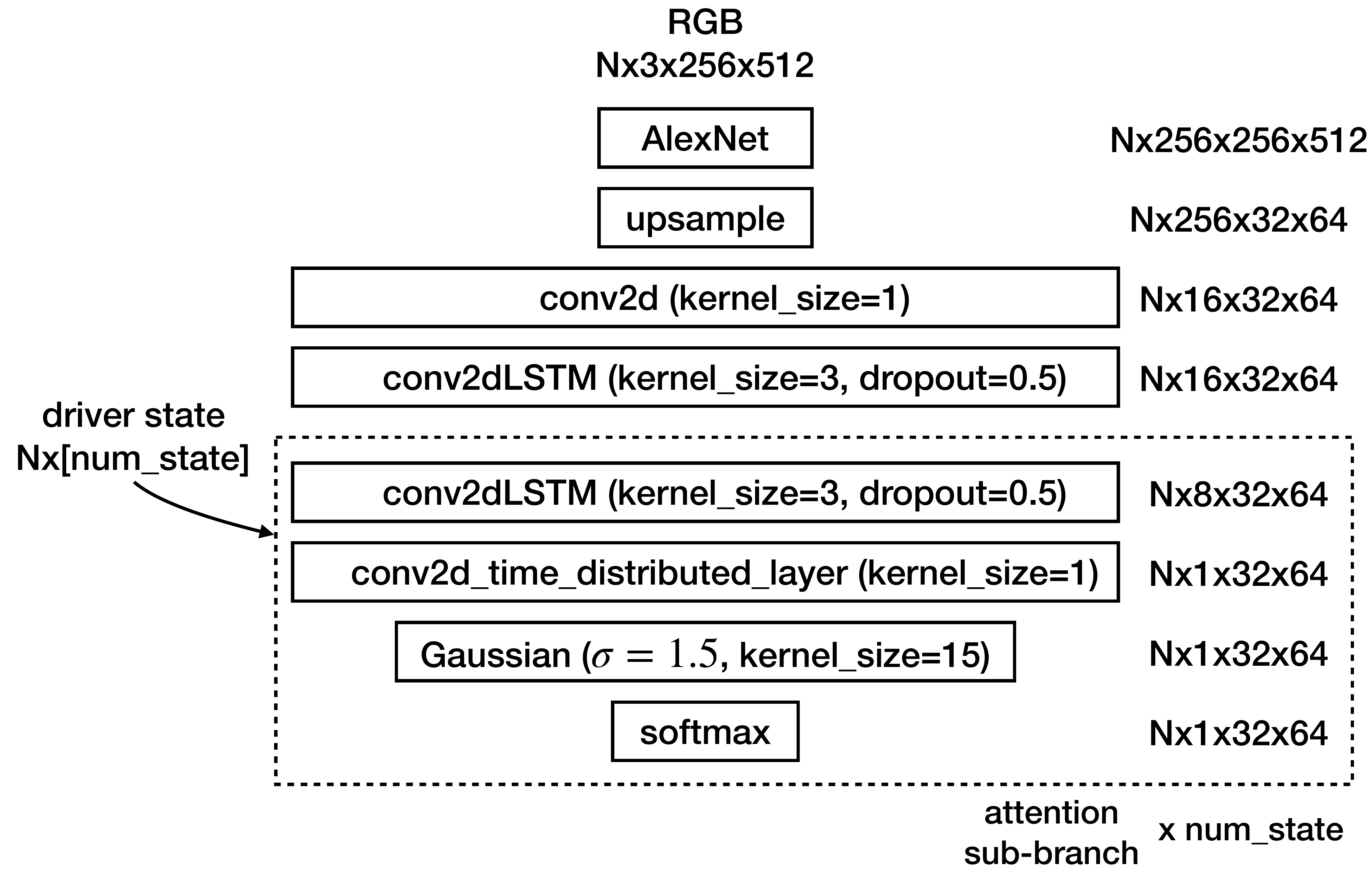}}
\caption{The architecture details of the BDD-A baseline and the multi-branch model. num\_state is 3 for the condition type of intersection intentions, and 2 for the condition type of distraction states. All of convolution layers are followed by batchnorm2d and ReLU piece-wise nonlinearity.}
\label{fig:multi_branch_baseline}
\end{figure}

For the standard convolution, all the input examples share the same kernel weights. However, for conditional convolution layer, i.e., CondConv~\cite{NEURIPS2019_f2201f51},  the kernel weights is input dependent. Condconv achieves this by keeping a linear combination of n kernel weights. We can interpret different kernel weights as different experts where each of these experts may perform better for one driver condition than the other. And then, a routing function generates weights that decide how much each expert contributes to the final convolution. Note that the original CondConv only takes feature maps from the previous layer as inputs. To have a driver-state dependent routing function, we modified on top of the CondConv by including 1D driver’s state as illustrated in Figure \ref{fig:modified_condconv}. 

\begin{figure}[t]
\centerline{\includegraphics[width=\columnwidth]{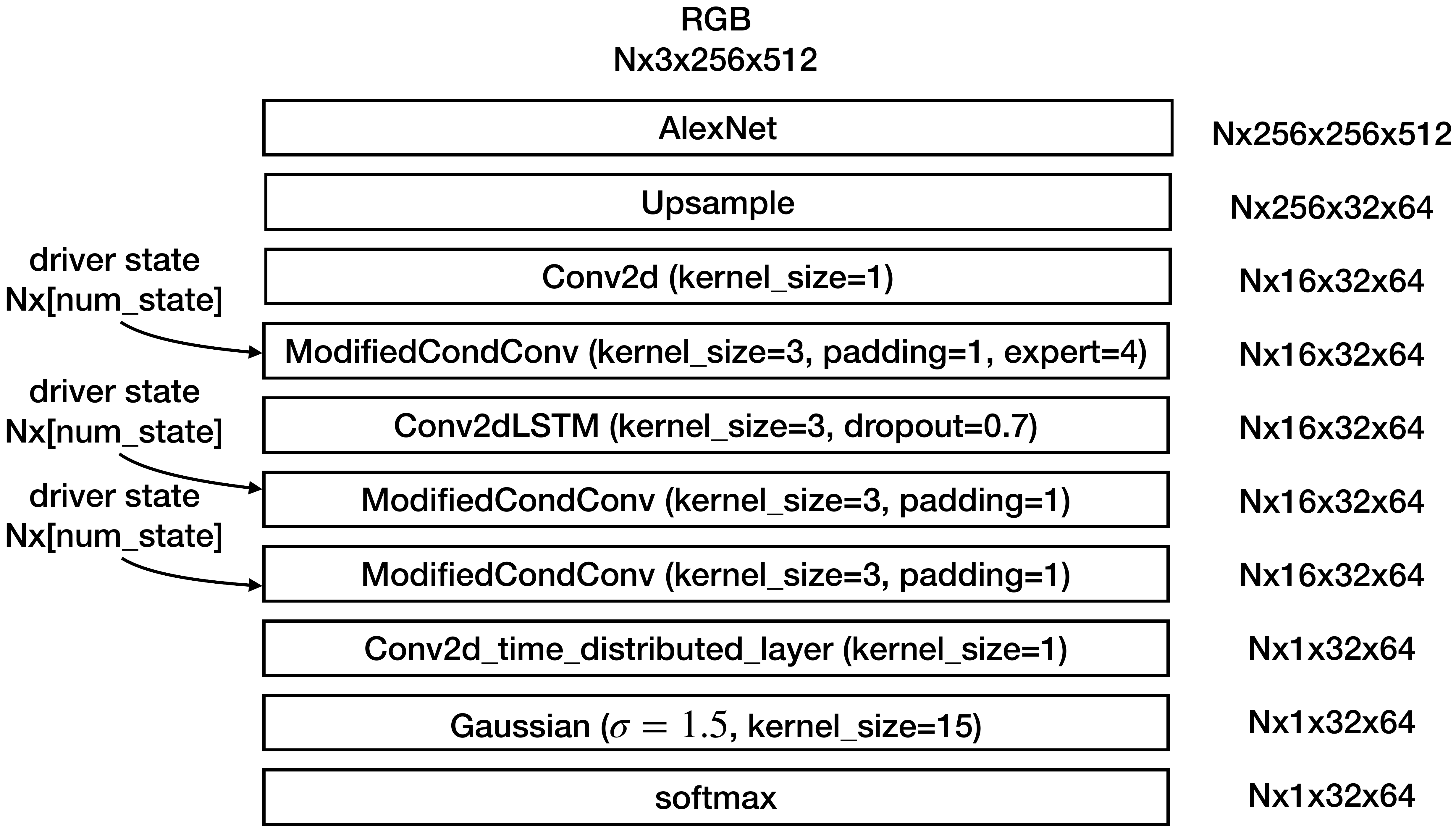}}
\caption{The architecture details of the modified condconv. num\_state is 3 for the condition type of intersection intentions, and 2 for the condition type of distraction states. All of convolution layers are followed by batchnorm2d and ReLU piece-wise nonlinearity. }
\label{fig:modificed_cond_conv_baseline}
\end{figure}

We follow the suggestion of the original author of CondConv to inject CondConv at multiple layers of our network and increase the dropout rate to reduce over-fitting from 0.5 to 0.7. The detail architecture is shown in Figure \ref{fig:modificed_cond_conv_baseline}. For simplicity, we only visualize one modified CondConv in Figure 4 of the main paper.

\begin{figure}[t]
\centerline{\includegraphics[width=\columnwidth]{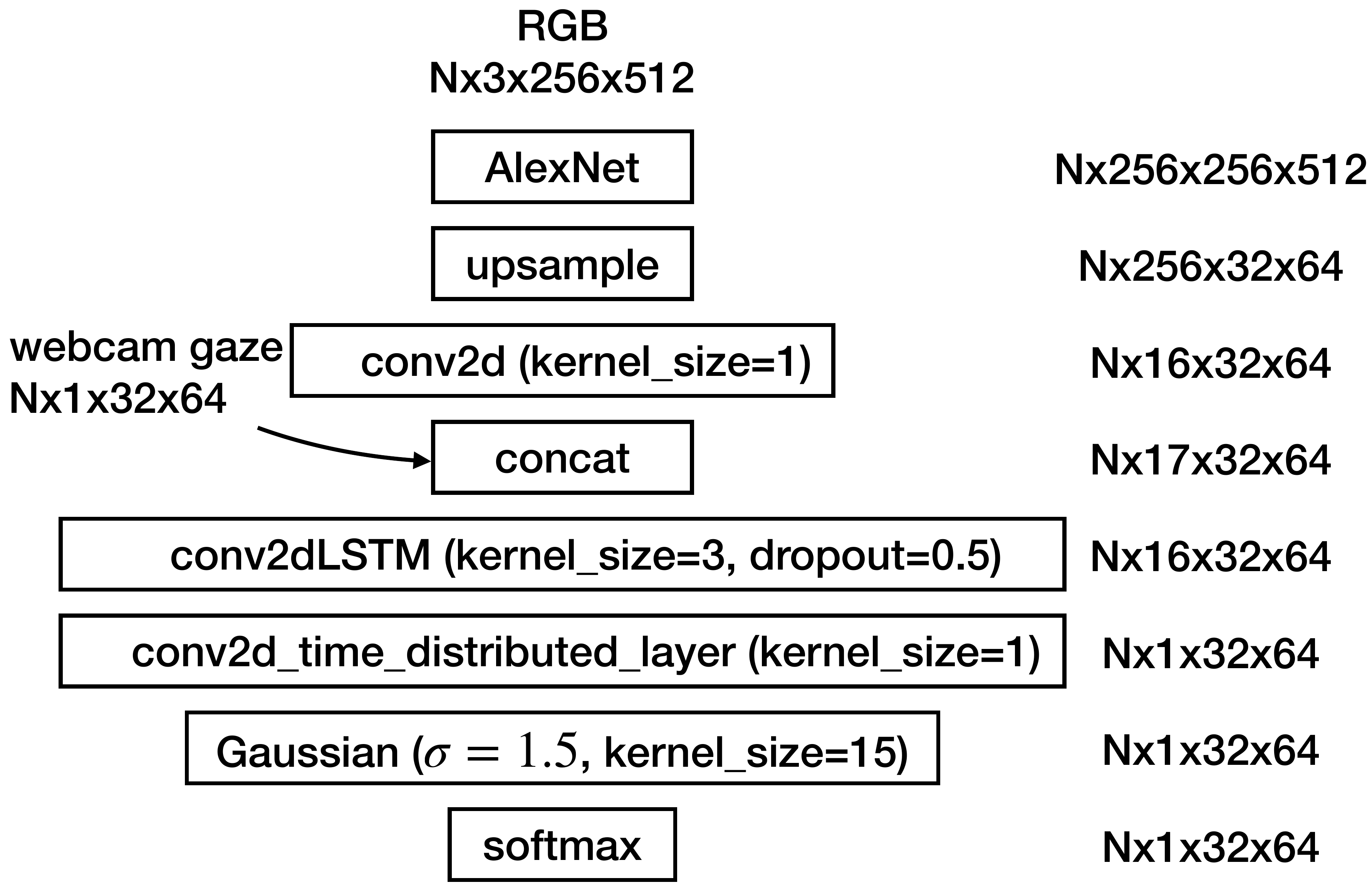}}
\caption{The architecture details of the gaze calibration network. All of convolution layers are followed by batchnorm2d and ReLU piece-wise nonlinearity. }
\label{fig:gaze_calibration_baseline}
\end{figure}

\subsection{Implementation Details for Gaze Calibration Network}
We visualize our architecture for the gaze calibration baseline in Figure \ref{fig:gaze_calibration_baseline}. The architecture is mostly the same as the unconditioned attention baseline, i.e., BDD-A. However, we concatenate the webcam gaze input with the encoded spatial feature from the upstream network.

\section{Experiments}
\begin{table*}[t]
\centering
    \caption{State-of-the-art model performance on our CoCAtt dataset.}
    \captionsetup{justification=centering}
    \label{table:no-intention-baseline-new}
    \renewcommand{\arraystretch}{1.0}
    \begin{tabular}{@{}llllllllll@{}} 
    \toprule 
      & & \multicolumn{4}{ c}{Autopilot Test Data} & \multicolumn{4}{ c}{Manual Test Data} \\
      \cmidrule(lr){3-6} \cmidrule(lr){7-10}
      Model & Training Data & \multicolumn{2}{ c}{Intersection}  & \multicolumn{2}{ c}{Lane-following} & \multicolumn{2}{ c}{Intersection}  & \multicolumn{2}{ c}{Lane-following} \\
       \cmidrule(lr){3-4} \cmidrule(lr){5-6} \cmidrule(lr){7-8} \cmidrule(lr){9-10}
       & & $D_{KL} \downarrow$ & $CC \uparrow$ & $D_{KL} \downarrow$ & $CC \uparrow$ & $D_{KL} \downarrow$ & $CC \uparrow$ & $D_{KL} \downarrow$ & $CC \uparrow$ \\
      \midrule
      \multirow{2}{*}{\small BDD-A} & autopilot & 2.07 & 0.36 & 
      1.46 & 0.47 & 
      1.78 & 0.43 & 
      1.47 & 0.50 \\
      & manual & 
      2.22 & 0.33 & 
      1.71 & 0.44 & 
      1.60 & 0.48 & 
      1.35 & 0.53 \\
      \multirow{2}{*}{\small DR(eye)VE} & autopilot & 
      2.20 & 0.40 & 
      1.90 & 0.45 & 
      1.74 & 0.50 & 
      1.77 & 0.47 \\
      & manual & 2.29 & 0.41 & 
      1.77 & 0.47 & 
      1.79 & 0.48 & 
      1.48 & 0.59 \\
      \bottomrule
    \end{tabular}
\end{table*}
\subsection{Evaluation Metrics}
\textbf{Prediction Entropy}
The prediction entropy~($H$) can characterize the confidence of the prediction for the cognitive-conditioned attention model. Specifically, we calculate the Prediction Entropy~($H$) as follows:
\begin{equation*}
\begin{split}
    H(P) = -\sum_i P_i \log P_i  \\
    \text{s.t.} \sum_i P_i = 1
\end{split}
\end{equation*}
, where P is one predicted attention map and i is the pixel index. 
\subsection{On the Effects of Driving Modes} \label{sec:exp1} 
The goal for this study is to report the performance of existing driver attention model on our dataset and reveal the learning difference between autopilot and manual drive attention data in CoCAtt dataset. 

We train and evaluate two state-of-the-art driver attention models, BDD-Attention and DR(eye)VE, on our dataset. The BDD-A model is re-implemented in PyTorch, and we reuse the single-stream implementation of the DR(eye)VE model with its pre-trained C3D encoder network as in ~\cite{pal2020looking}. Table~\ref{table:no-intention-baseline-new} shows a comparison of the BDD-A and DR(eye)VE models trained under both autopilot and manual modes with no intention or distraction conditioning. Each model is evaluated on both the autopilot and manual test sets. 

As shown in Table~\ref{table:no-intention-baseline-new}, both models exhibit worse performance if trained and tested on the autopilot attention data, indicating that autopilot attention has more variability and is thus harder to model if no planned action is known beforehand. Additionally, since existing driver attention datasets contain only manual drive data, we explore the transfer capability of models trained in one setting and applied in another. We observe that models trained on autopilot data can transfer better to the manual drive setting than manual-to-autopilot. Our intuition is that drivers in autopilot mode are more likely to attend to as many potential driving cues as possible to account for not knowing the planned action of the ego car in advance. 

\begin{figure}[ht]
\centerline{\includegraphics[width=0.45\textwidth]{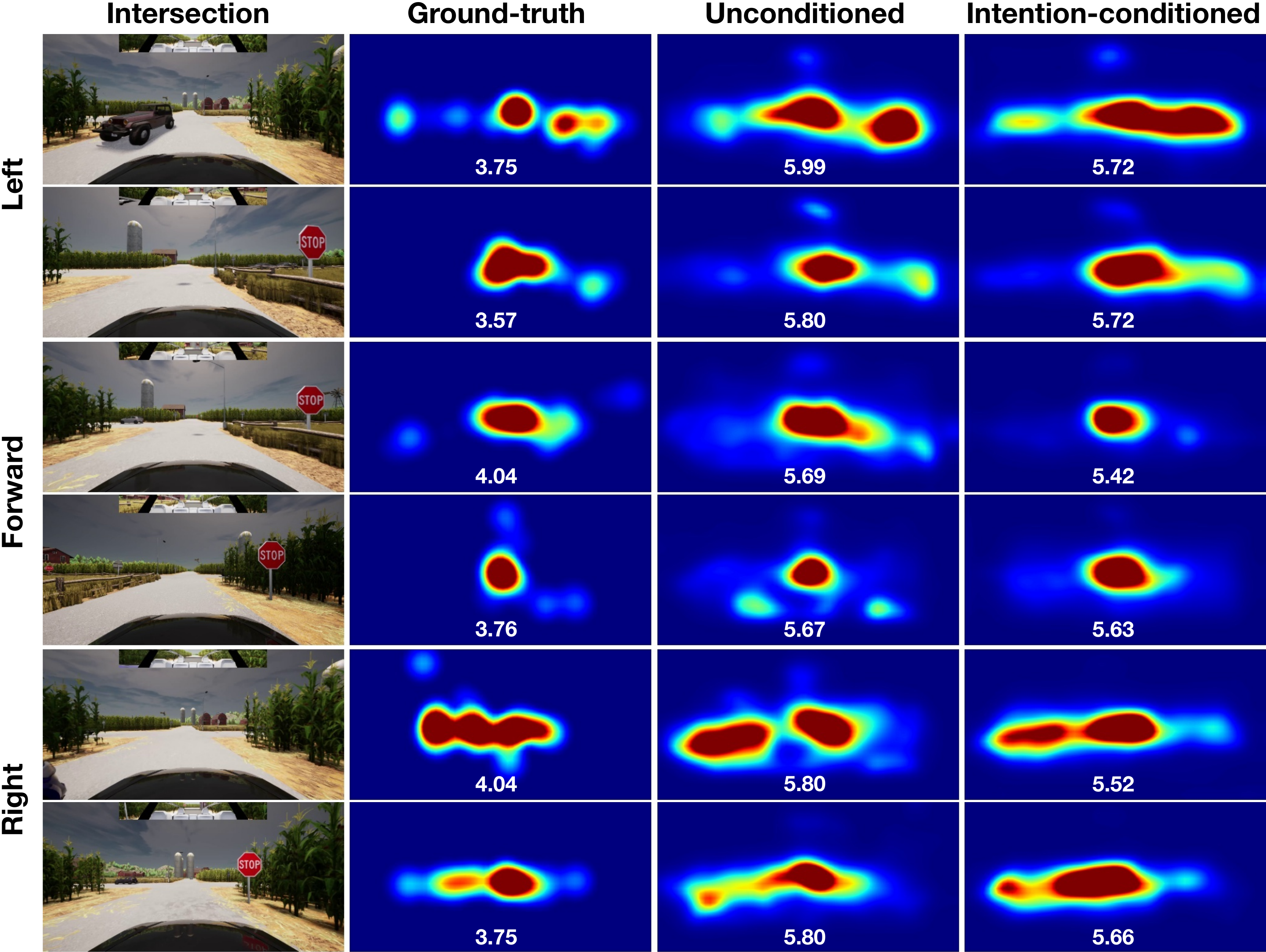}}
\caption{Qualitative results for the gaze prediction with and without intention conditioning. The last three columns are cumulative attention predictions for the scene shown in the leftmost column. Each intention (labeled on left) has two rows of examples. The white numbers represent the entropy of predicted attention maps.\label{fig:exp2_qualitative_res}}
\end{figure}

\subsection{Qualitative Study On the Effects of Driver States}
In Section 6.2 of our paper, our quantitative results suggest that cognitive-conditioned learning can improve driver attention prediction. The above findings are validated by our qualitative results as well as shown in Figure \ref{fig:exp2_qualitative_res}. We use the multi-branch baseline for the qualitative study due to its flexibility to add attention to sub-branchs for different driver states without re-training the backbone network. Note that the same conclusion can be reached from the modified CondConv baseline for the qualitative study. As human attention is noisy for individual frames~\cite{dreyeve, pal2020looking}, we choose to aggregate attention maps over a long time horizon (e.g., intervals when approaching intersections), to compare the ground-truth gazes and both unconditioned and conditioned gaze predictions. Each cumulative attention heatmap is generated by aggregating gaze maps when the driver is approaching an intersection. In Figure~\ref{fig:exp2_qualitative_res}, we visualize six cumulative attention heatmaps for the intention-conditioned case at a four-way intersection scenario with three different intentions. Compared to the unconditioned case, the intention-conditioned model generates more focused attention with fewer weights allocated in regions unrelated to the current driver intention.

\section{Application\label{sec:application}}

Aside from performance improvement, our cognitive-conditioned attention model can solve new real-world challenges.
To demonstrate its practicality, we showcase one application of our cognitive-conditioned attention model on road safety analysis. This task aims to assess the safety standard of roads to reduce the chance of accidents~\cite{KANUGANTI20174649, inproceedings_rsa}. Traditional road safety analysis uses survey-based methods or historical data on a case-by-case basis~\cite{inproceedings_perceived_risk, KANUGANTI20174649, Machsus_2017, safety6040045}. However, our cognitive-conditioned attention model offers an alternative, data-driven solution that can be generalized to address diverse road patterns. Our approach is built on the findings of previous cognitive distraction studies related to the positive correlation between cognitively distracted driving and risk of car accidents~\cite{cdc_distracted_driving, doi:10.1177/1541931218621441}. To estimate the risk of car accidents per road segment, we measure the gaze behavior difference between distracted and attentive drivers. The more significant the difference, the more likely the distracted driver will miss important driving cues. We quantify gaze behavior difference with the Earth Mover's Distance, which has the advantage over other metrics of measuring the spatial distance cost to migrate from distracted to attentive attention maps~\cite{bylinskii2018different}. 

\begin{figure}[t]
\centerline{\includegraphics[width=\columnwidth]{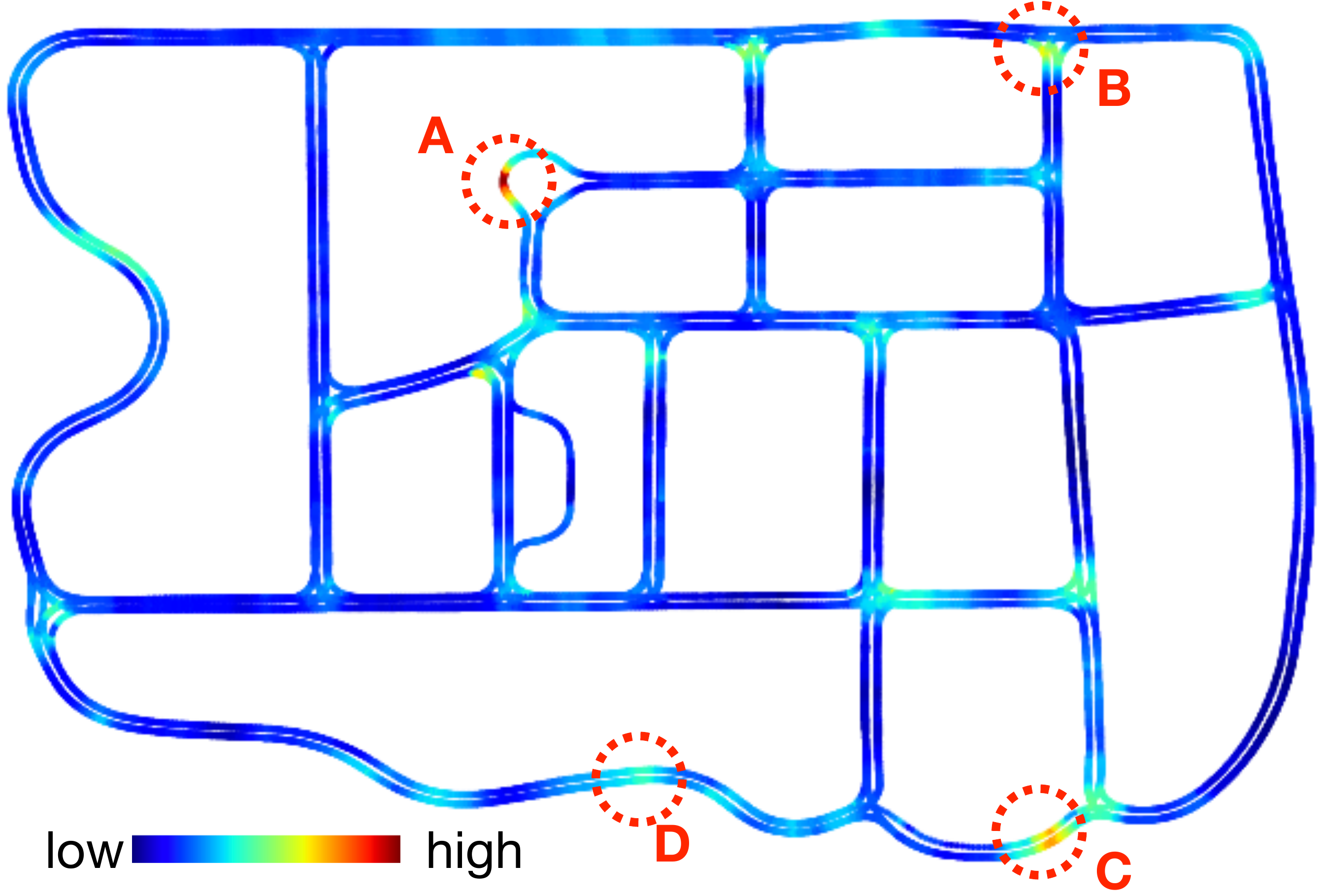}}
\caption{Visualization for one use case of our cognitive-conditioned attention model on road safety analysis. The color indicates the distraction risk, which we estimate by computing the difference in gaze behavior between distracted and attentive drivers.}
\label{fig:use_case}
\end{figure}

We re-trained our distraction-state-conditioned model over the autopilot data. We then predicted both attentive and distracted driver attention maps for each timestamp by feeding attentive and distracted cognitive states as two separate inputs into our network. Figure~\ref{fig:use_case} shows the visualization of our result after applying a median filter over the local neighborhood. Our approach offers location-wise distraction risk measurements. For example, our method estimates high risk at the round-about (point A in Figure~\ref{fig:use_case}).
This insight may suggest that traffic warning signs here would be useful to notify drivers to look for cars as they enter the round-about. Another finding is that road segments approaching and exiting traffic intersections are more accident-prone (point B and C). While most of the highlighted regions are near intersection areas, our proposed approach can also detect risky areas of other road conditions. For example, the road region in point D, which has a steep slope increase, has a higher value than nearby regions.
This proposed analysis is a new method for determining driver risk across roadways and driving scenarios. Such data-driven insights exhibit the practicality and applicability of our cognitive-conditioned attention model for road safety analysis and other potential real-world applications.

\bibliographystyle{IEEEtran}
\bibliography{proceedings}
\end{document}